\documentclass[conference]{IEEEtran}

\usepackage{graphics}
\usepackage{graphicx}
\usepackage{amsmath}
\usepackage{amssymb}
\usepackage{booktabs}
\usepackage{multirow}
\usepackage{xcolor}
\usepackage{hyperref}
\usepackage{placeins}
\usepackage{stfloats}
\usepackage{capt-of}

\definecolor{clrOriginal}{RGB}{76,175,80}
\definecolor{clrTarget}{RGB}{66,133,244}
\definecolor{clrObject}{RGB}{255,152,0}
\definecolor{clrInteract}{RGB}{156,39,176}

\title{\LARGE \bf
LangGap: Diagnosing and Closing the Language Gap\\in Vision-Language-Action Models\thanks{Code and benchmark will be available at: \url{https://github.com/YC11Hou/langgap}}
}

\author{
\IEEEauthorblockN{Yuchen Hou and Lin Zhao}\\
\IEEEauthorblockA{
Department of Electrical and Computer Engineering, National University of Singapore, Singapore\\
Email: e1520673@u.nus.edu, zhaolin@nus.edu.sg
}
}

\begin{document}

\maketitle
\thispagestyle{empty}
\pagestyle{empty}

\begin{abstract}
Vision-Language-Action (VLA) models achieve over 95\% success on standard benchmarks. However, through systematic experiments, we find that current state-of-the-art VLA models largely ignore language instructions. Prior work lacks: (1)~systematic semantic perturbation diagnostics, (2)~a benchmark that forces language understanding by design, and (3)~linguistically diverse training data.

This paper constructs the \textbf{LangGap benchmark}, based on a four-dimensional semantic perturbation method---varying instruction semantics while keeping the tabletop layout fixed---revealing language understanding deficits in $\pi$0.5. Existing benchmarks like LIBERO assign only one task per layout, underutilizing available objects and target locations; LangGap fully diversifies pick-and-place tasks under identical layouts, forcing models to truly understand language.

Experiments show that targeted data augmentation can partially close the language gap---success rate improves from 0\% to 90\% with single-task training, and 0\% to 28\% with multi-task training. However, as semantic diversity of extended tasks increases, model learning capacity proves severely insufficient; even trained tasks perform poorly. This reveals a \textbf{fundamental challenge for VLA models in understanding diverse language instructions}---precisely the long-term value of LangGap.
\end{abstract}

\section{Introduction}

Vision-Language-Action (VLA) models represent a significant paradigm shift in robotic manipulation---unifying visual perception, language understanding, and action generation within a single end-to-end architecture. These models have achieved remarkable success on standard benchmarks, with state-of-the-art models such as $\pi$0.5 reaching over 95\% success rates on LIBERO~\cite{liu2023libero}. However, growing evidence suggests that these models largely ignore language instructions, instead relying on visual shortcuts. LIBERO-PRO~\cite{zhou2025liberopro} finds that models produce identical trajectories even when given meaningless text; LIBERO-Plus~\cite{fei2025liberoplus} tests seven perturbation types (including surface-level changes like paraphrasing) and finds that language perturbations cause the second smallest performance drop, indicating models simply do not attend to language; BayesVLA~\cite{ke2025bayesvla} formalizes this as ``modality imbalance''---insufficient language diversity in training data causes models to learn to ignore linguistic signals.

While the problem has been established, existing diagnostic and solution approaches remain inadequate. At the \textbf{diagnostic level}, prior work either conducts only surface-level paraphrase tests or provides coarse conclusions like ``language is ignored'', lacking fine-grained analysis of which specific semantic dimensions fail---we do not know whether models fail to understand object nouns, target locations, or spatial relations. At the \textbf{benchmark level}, existing benchmarks like LIBERO typically assign only one task per tabletop layout, allowing models to complete tasks through visual memorization alone without truly understanding language---the same visual input always corresponds to the same action output. At the \textbf{training level}, existing solutions primarily focus on architectural modifications (e.g., rebalancing modality weights) rather than addressing the root cause of insufficient language diversity at the data level.

This paper systematically addresses these issues from three perspectives: diagnosis, benchmark, and training. Our main contributions are as follows:

\begin{itemize}
    \item \textbf{Diagnostic method}: We propose a four-dimensional semantic perturbation taxonomy that varies instruction semantics along object category, target location, spatial description, and interactive action dimensions while keeping the tabletop layout fixed. This reveals differential failure modes in $\pi$0.5---Change Target yields 0\% success while Change Object achieves 29.3\%. Such fine-grained analysis cannot be provided by prior coarse-grained conclusions.
    \item \textbf{Evaluation benchmark}: We construct the LangGap benchmark (99 tasks), which fully utilizes manipulable objects and target locations in each tabletop layout, ensuring diverse instructions under identical visual inputs and forcing models to truly understand language. This is the first VLA evaluation benchmark that enforces language reliance by design.
    \item \textbf{Training validation}: Through progressive experiments, we validate that targeted data augmentation can partially close the language gap (single-task improves from 0\% to 90\%), but as task scale increases, learning difficulty rises sharply, revealing the fundamental challenge for VLA models in understanding diverse language.
\end{itemize}

Unlike existing benchmarks that reach performance saturation within months, LangGap's same-scene diverse semantic task design ensures long-term evaluation value for future VLA development.

\section{Related Work}

\subsection{Vision-Language-Action Models}

VLA models have developed rapidly in robotic manipulation in recent years. \textbf{RT-1} and \textbf{RT-2}~\cite{brohan2022rt1,brohan2023rt2} are early representative works that introduced Transformer architectures to robot control, demonstrating the effectiveness of large-scale data training. The \textbf{$\pi$0} series~\cite{black2024pi0} adopts flow matching methods, achieving over 95\% success rates on benchmarks like LIBERO. Open-source models such as \textbf{OpenVLA}~\cite{kim2024openvla} and \textbf{SmolVLA}~\cite{allauzen2025smolvla} have lowered research barriers and promoted community development. Additionally, modular approaches like \textbf{CLIPort}~\cite{shridhar2022cliport} and \textbf{SayCan}~\cite{ahn2022saycan} decouple vision-language models from robot policies, exploring an alternative technical route, along with language-conditioned manipulation methods~\cite{lynch2021language}. Other notable approaches include generalist robot policies such as Octo~\cite{octo2024} and RoboFlamingo~\cite{li2023roboflamingo}, as well as multi-task transformers like Perceiver-Actor~\cite{shridhar2023perceiver}. These models build on vision-language pretraining foundations~\cite{radford2021clip,driess2023palme}. Despite excellent performance on standard benchmarks, multiple studies indicate fundamental limitations in language understanding.

\subsection{Manipulation Benchmarks and Robustness Evaluation}

\textbf{LIBERO}~\cite{liu2023libero} is a widely-used robotic manipulation benchmark containing multiple task suites (Spatial, Object, Goal, etc.) for evaluating multi-task learning capabilities. Building on this, \textbf{LIBERO-Plus}~\cite{fei2025liberoplus} proposes robustness tests with seven perturbation types, finding that language perturbations cause relatively small performance drops ($-$25.3\%), and provides 20,000+ trajectories of environmental variation training data. \textbf{LIBERO-PRO}~\cite{zhou2025liberopro} focuses on diagnostic evaluation, testing perturbations across object, position, instruction, and environment dimensions, revealing model fragility under task perturbations. Beyond LIBERO, other manipulation benchmarks such as RLBench~\cite{james2020rlbench}, Meta-World~\cite{yu2020metaworld}, and CALVIN~\cite{mees2022calvin} also evaluate multi-task generalization, though none enforce language reliance by design. These works provide important foundations for evaluating VLA model robustness.

\subsection{Modality Imbalance and Architectural Solutions}

Another research line focuses on why language is ignored and potential solutions. \textbf{BayesVLA}~\cite{ke2025bayesvla} formalizes this as a modality imbalance problem, \textbf{LangForce}~\cite{wang2026langforce} identifies information collapse, and Residual Semantic Steering~\cite{chen2026rss} diagnoses modality collapse. All three propose architectural modifications to rebalance modalities. Complementing these architectural analyses, Wanna et al.~\cite{rosenman2026linguisticdiversity} audit VLA training datasets and find that fewer than 2\% of instructions are linguistically unique, providing a data-level explanation for why models learn to ignore language.

\subsection{Distinction from Prior Work}

Table~\ref{tab:related_comparison} summarizes the key distinctions between our work and prior approaches.

\begin{table}[ht]
\centering
\caption{Comparison with related work on VLA language understanding}
\label{tab:related_comparison}
\footnotesize
\begin{tabular}{lcccc}
\toprule
 & \rotatebox{60}{LIBERO-Plus} & \rotatebox{60}{LIBERO-PRO} & \rotatebox{60}{Arch.$^\dagger$} & \rotatebox{60}{Ours} \\
\midrule
Perturbation type & Surface & General & -- & Semantic \\
Failure taxonomy & $\times$ & $\times$ & $\times$ & \checkmark \\
Same-scene design & $\times$ & $\times$ & $\times$ & \checkmark \\
Training data & Env. & None & None & Lang. \\
Training validation & $\times$ & $\times$ & -- & \checkmark \\
Approach & Bench. & Bench. & Arch. & Data+Bench. \\
\bottomrule
\end{tabular}
\vspace{2pt}

{\scriptsize $^\dagger$BayesVLA~\cite{ke2025bayesvla}, LangForce~\cite{wang2026langforce}, Zhan et al.~\cite{chen2026rss}}
\end{table}

\section{Method}

\subsection{Semantic Perturbation Framework}

We design a systematic evaluation framework to diagnose \textit{which specific semantic dimensions} VLA models fail on. The core design principle is \textbf{maximal semantic diversity within identical visual layouts}: for each LIBERO~\cite{liu2023libero} scene, we identify all manipulable object categories, reachable target locations, spatial descriptions to distinguish different instances of the same manipulated object category, and valid action types (pick-and-place, drawer open/close). We generate every physically valid semantic combination, designing multiple distinct tasks for each scene. These are genuinely \textit{different tasks}---not paraphrases---that share the same initial visual state, forcing models to demonstrate language understanding rather than visual memorization.

We organize these semantic variations into four orthogonal perturbation dimensions, each targeting a distinct compositional element of manipulation instructions~\cite{lake2018compositional}:

\begin{itemize}
    \item \textbf{Change Object Category}: Change the category of the manipulated object (e.g., ``pick \textit{bowl} $\rightarrow$ ``pick \textit{ramekin}). Tests understanding of object category nouns.
    \item \textbf{Change Target}: Change the target location (e.g., ``place on \textit{plate} $\rightarrow$ ``place on \textit{stove}). Tests understanding of spatial goals.
    \item \textbf{Spatial Description}: For different instances of the same manipulated object category, distinguish them by changing spatial descriptions (e.g., ``the bowl to the right of the \textit{ramekin} $\rightarrow$ ``the bowl to the right of the \textit{plate}). Tests understanding of spatial relationships.
    \item \textbf{Drawer Action}: Change the action type (e.g., ``\textit{put} bowl $\rightarrow$ ``\textit{open} drawer). Tests understanding of action semantics.
\end{itemize}

\textbf{Diagnostic principle}: A model with genuine compositional understanding should maintain reasonable success rates across all dimensions. Results on each dimension reveal the model's understanding of the corresponding component in complete language instructions. Compared to the coarse conclusion of ``language is ignored in prior work, this paper provides more fine-grained diagnosis.

\begin{figure}[ht]
\centering
\includegraphics[width=0.90\columnwidth]{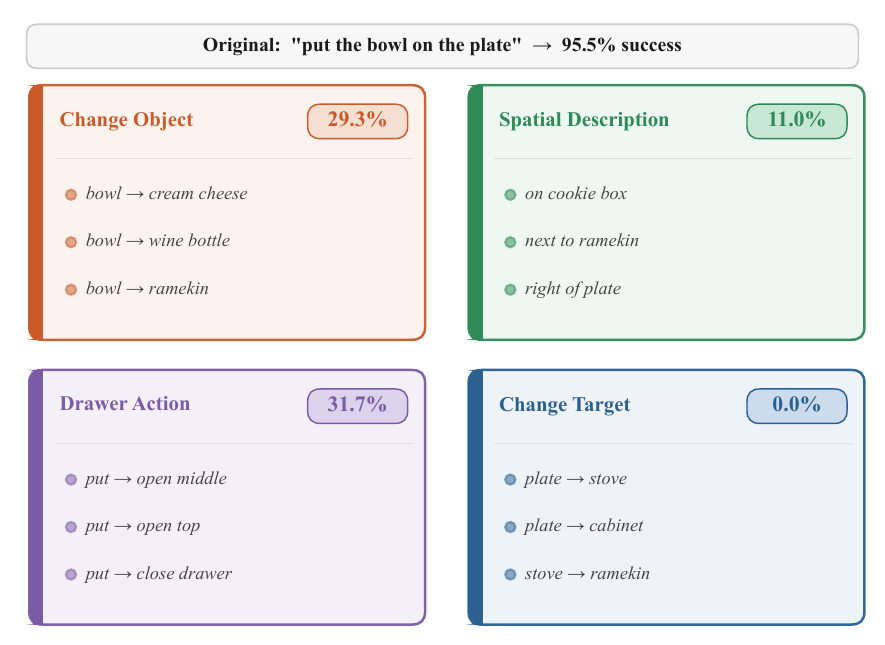}
\caption{Four semantic perturbation types with example changes and observed success rates.}
\label{fig:perturbation_types}
\end{figure}

We leverage LIBERO benchmark suites, selecting original official tasks suitable for semantic variation and extension of instructions. Table~\ref{tab:suite_dimensions} shows which perturbation dimensions can be designed for each suite.

\begin{table}[ht]
\centering
\caption{Perturbation dimensions tested in each LIBERO suite}
\label{tab:suite_dimensions}
\begin{tabular}{lcccc}
\toprule
Suite & Ch.\ Obj. & Ch.\ Tgt. & Sp.\ Desc. & Drawer \\
\midrule
libero\_spatial & \checkmark & \checkmark & \checkmark & \checkmark \\
libero\_object & \checkmark & $\times$ & $\times$ & $\times$ \\
libero\_goal & \checkmark & \checkmark & $\times$ & $\times$ \\
\bottomrule
\end{tabular}
\end{table}

\textbf{Evaluation protocol}: Each task is evaluated for 20 episodes with binary success. We report per-task, per-dimension, and per-suite breakdowns.

\subsection{Diagnostic Evaluation}

We apply the evaluation framework to diagnose semantic understanding in state-of-the-art VLA models.

\textbf{Setup}: We evaluated $\pi$0.5 on 99 tasks (40 original LIBERO tasks + 59 extended perturbation tasks), totaling 1980 episodes. Each task was evaluated for 20 episodes.

\begin{figure}[ht]
\centering
\includegraphics[width=0.75\columnwidth]{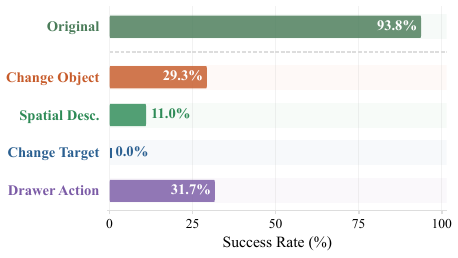}
\caption{VLA performance under semantic perturbations. Original tasks achieve over 95\% success on several suites. Semantic changes cause severe failures: Change Object 29.3\%, Spatial Description 11.0\%, Drawer Action 31.7\%, and Change Target \textbf{0\%} across all 13 tasks.}
\label{fig:perturbation_results}
\end{figure}

\textbf{Results}: Results are shown in Figure~\ref{fig:perturbation_results} and Table~\ref{tab:main_results}.

\begin{table}[ht]
\centering
\caption{Diagnostic evaluation results on $\pi$0.5}
\label{tab:main_results}
\begin{tabular}{lcccc}
\toprule
Category & Tasks & Episodes & Success & Rate \\
\midrule
\multicolumn{5}{l}{\textit{By data source}} \\
Original (LIBERO) & 40 & 800 & 750 & 93.8\% \\
Extended (Ours) & 59 & 1180 & 253 & 21.4\% \\
Total & 99 & 1980 & 1003 & 50.7\% \\
\midrule
\multicolumn{5}{l}{\textit{By perturbation dimension}} \\
Change Object & 38 & 760 & 223 & 29.3\% \\
Change Target & 13 & 260 & 0 & 0.0\% \\
Spatial Description & 5 & 100 & 11 & 11.0\% \\
Drawer Action & 3 & 60 & 19 & 31.7\% \\
\bottomrule
\end{tabular}
\end{table}

\textbf{Performance Gap}: Original 93.8\% $\rightarrow$ Semantic perturbations 21.4\% (\textbf{$-$72.4\%})

\textbf{Analysis}:

\textbf{Finding 1: Complete failure on target changes (0\%)}

\begin{figure*}[t]
\centering
\includegraphics[width=0.95\textwidth]{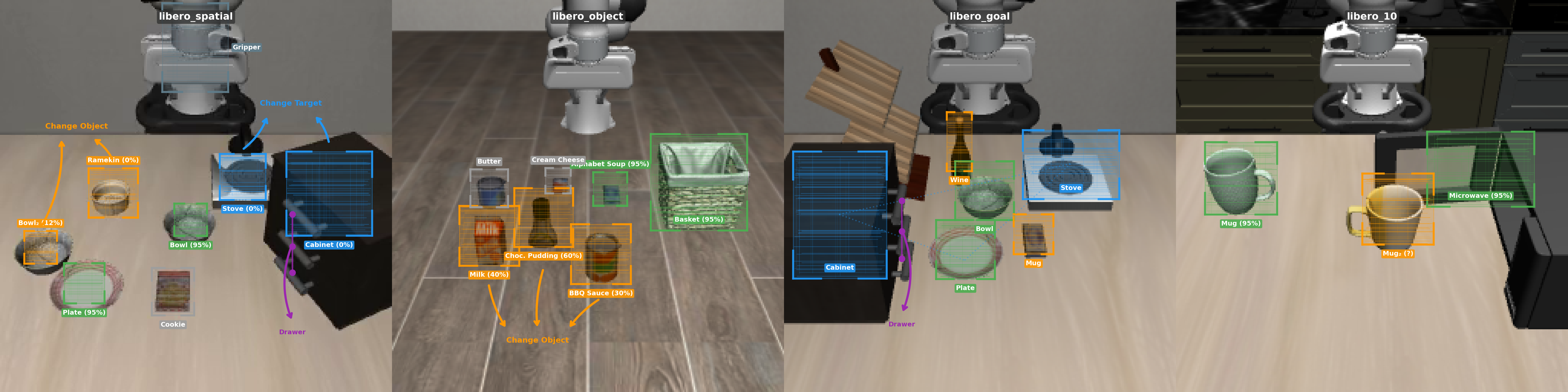}\\[4pt]
{\footnotesize
\textcolor{clrOriginal}{\rule[3pt]{12pt}{2pt}}\;\,Original object/target\hspace{14pt}
\textcolor{clrTarget}{\rule[3pt]{12pt}{2pt}}\;\,Alternative target\hspace{14pt}
\textcolor{clrObject}{\rule[3pt]{12pt}{2pt}}\;\,Alternative object\hspace{14pt}
\textcolor{clrInteract}{\rule[3pt]{12pt}{2pt}}\;\,Interaction point}
\caption{Evaluation scenes from LIBERO suites that support multiple semantic variations.}
\label{fig:scene_annotated}
\end{figure*}

Our taxonomy isolates a failure mode that prior work's coarser evaluations could not pinpoint: across all 13 change-target tasks (260 episodes), the success rate is exactly 0\%. While prior work has shown aggregate language neglect, our per-dimension breakdown reveals that the \textit{spatial goal slot} is the most severely ignored compositional element---the model never redirects to an alternative target location, even when the manipulation object and action type remain unchanged. Figure~\ref{fig:scene_annotated} illustrates the evaluation scenes with annotated semantic variation elements.

\textbf{Finding 2: Non-uniform failure across dimensions reveals differential understanding}

The success rate of the Change Object dimension ranges from 0\% to 80\% across different tasks, in stark contrast to the Change Target dimension where success rate is consistently 0\%. This non-uniformity can only be observed through our per-dimension taxonomy, revealing that the model has partial understanding of object categories (succeeding when the modified instruction happens to match training patterns). Prior work's results obscured the differences in model understanding across different semantic dimensions.

\subsection{LangGap Benchmark}

Based on the differential failure modes revealed by our diagnostic experiments across different semantic perturbation dimensions, we construct LangGap to systematically evaluate and improve VLA language understanding. LangGap is designed to \textit{force} language reliance by construction.

\textbf{Design Principles}:

\textbf{(1) Same-scene multi-task}: Multiple tasks share identical initial visual states, completely eliminating visual shortcuts. A model that ignores language cannot distinguish between tasks in the same scene and will achieve at most $1/k$ success rate (where $k$ is the number of tasks per scene), making language the \textit{only} available discriminative signal.

\textbf{(2) Instruction-level split}: Training tasks do not include all test tasks, ensuring that test evaluation contains language instructions not seen during training.

\textbf{(3) Physical feasibility validation}: All extended tasks are verified in the LIBERO simulator~\cite{liu2023libero} to ensure that the target object is graspable, the placement location is reachable, and the success condition is detectable. This eliminates confounds from impossible tasks.

\textbf{Benchmark Statistics}: After physical feasibility checks, we obtain 59 valid extended tasks across 3 suites:
\begin{itemize}
    \item libero\_spatial: 28 tasks (covering all four dimensions)
    \item libero\_goal: 9 tasks (Change Object Category, Change Target)
    \item libero\_object: 22 tasks (Change Object Category)
\end{itemize}

Each extended task shares the \textit{identical} initial scene state as its source original task, ensuring that only the language instruction differs.

\begin{table}[ht]
\centering
\caption{LangGap benchmark statistics}
\label{tab:benchmark_stats}
\scriptsize
\begin{tabular}{lccc}
\toprule
 & LIBERO & LIBERO-Plus & LangGap \\
\midrule
Total tasks & 130 & 10{,}030 & 99 \\
Extended (semantic) & 0 & -- & 59 \\
Same-scene tasks & 10 & -- & 59 \\
Ext.\ w/ training data & 0 & -- & 16 \\
Training trajectories & 3{,}000 & 20{,}000+ & $\sim$4{,}100 \\
Perturbation dims. & -- & 7$^\dagger$ & 4 \\
Semantic focus & Surface & Mixed & Compositional \\
\bottomrule
\end{tabular}
\vspace{2pt}

{\scriptsize $^\dagger$LIBERO-Plus perturbs visual and language dimensions; only 1/7 is linguistic.}
\end{table}

\subsection{Training Data Collection}

\textbf{Data split principle}: We adopt instruction-level split. From the 59 extended tasks, we select 16 representative tasks across three suites as the training set:
\begin{itemize}
    \item libero\_spatial: 9 tasks
    \item libero\_goal: 3 tasks
    \item libero\_object: 4 tasks
\end{itemize}

These 16 tasks cover all four perturbation dimensions. The remaining 43 extended tasks serve as the held-out test set, sharing the same visual scenes as training tasks but using different semantic instructions.

\textbf{Collection method}: We use waypoint-based controllers with OSC\_POSE control in robosuite~\cite{zhu2020robosuite}, collecting scripted demonstrations following the behavioral cloning paradigm~\cite{pomerleau1991alvinn,mandlekar2021robomimic}. We collect approximately 150 demonstrations per task, each with standard environment initialization, yielding approximately 2,400 training episodes.

\section{Experiments}

We validate that the language understanding gap identified by our taxonomy can be narrowed through targeted multi-task training data, without architectural modifications.

\subsection{Setup}

We construct five training configurations at increasing scale (Table~\ref{tab:setup}). All models fine-tune $\pi$0.5~\cite{black2024pi0} with LoRA~\cite{hu2022lora} on a single RTX~4090 GPU (learning rate $1\times10^{-4}$, batch size~8). Data collection follows the scripted waypoint approach described in Section~3.4.

Experiments are designed as two paired comparisons. Experiments~2 and~3 form one pair, while Experiments~4 and~5 form another. Both pairs compare training with extended data only versus extended data plus official data. The difference is that Experiments~2 and~3 use extended data from a single suite (libero\_spatial), configured as 6-task and 45-task respectively; Experiments~4 and~5 use extended data across three suites (16 extended tasks), configured as 16-task and 56-task respectively.

Success rate is the primary metric: binary per-episode, with per-task and per-dimension breakdowns.

\begin{table}[ht]
\centering
\caption{Experimental setup across five configurations}
\label{tab:setup}
\scriptsize
\begin{tabular}{lcccccc}
\toprule
 & Tasks & Demos & $r$ & Steps & Eval Tasks & Ep \\
\midrule
Exp~1 & 1 ext & 56 & 32 & 20K & 1 & 10 \\
Exp~2 & 1+5 ext & 300 & 64 & 200K & 5 ext & 10 \\
Exp~3 & 40+5 ext & $\sim$2{,}000 & 64 & 200K & 5 ext & 10 \\
Exp~4 & 16 ext & $\sim$2{,}400 & 64 & 200K & 16 ext & 5 \\
Exp~5 & 40+16 ext & 4{,}093 & 64 & 200K & 16 ext & 5 \\
\bottomrule
\end{tabular}
\end{table}

\subsection{Results}

\setcounter{dbltopnumber}{1}
\begin{table*}[!t]
\centering
\small
\caption{Progressive validation across scales. Baseline refers to pretrained $\pi$0.5 evaluated on corresponding tasks.}
\label{tab:progressive}
\begin{tabular}{llccc}
\toprule
Experiment & Training Config & Eval Tasks & Baseline & Ours \\
\midrule
Single-task & 1 ext & 1 & 3.75 & \textbf{90.0} \\
\midrule
\multicolumn{5}{l}{\textit{Spatial focused (5 extended tasks from libero\_spatial)}} \\
6-task & 1 orig + 5 ext & 5 ext & 0.0 & \textbf{28.0} \\
45-task & 40 orig + 5 ext & 5 ext & 0.0 & 4.0 \\
\midrule
\multicolumn{5}{l}{\textit{Multi-suite (16 extended tasks across 3 suites)}} \\
16-task & 16 ext only & 16 ext & 26.2 & 6.2 \\
56-task & 40 orig + 16 ext & 16 ext & 26.2 & 27.5 \\
\bottomrule
\end{tabular}
\end{table*}

Table~\ref{tab:progressive} presents our central findings across training scales. Single-task fine-tuning achieves 90\% from 3.75\%.

For libero\_spatial training, the 6-task setting reaches 28\% on extended tasks; when combined with official tasks to form 45-task, the success rate is diluted to 4\%.

For multi-suite training, 16-task (extended only) achieves 6.2\%, while 56-task (with official tasks) reaches 27.5\% overall but only 6.7\% on spatial tasks. This indicates that official data is effective on extended tasks similar to it, but has no effect or even degrades performance on other extended tasks.

\textbf{Full benchmark evaluation.} Table~\ref{tab:benchmark_full} presents comprehensive evaluation results of state-of-the-art models on our benchmark, covering 40 official LIBERO tasks and 59 extended tasks (99 total), with per-suite breakdowns. Table~\ref{tab:benchmark_dimension} provides detailed results on extended tasks, broken down by perturbation dimension.

\FloatBarrier
\setcounter{dbltopnumber}{2}

\begin{table*}[ht]
\centering
\caption{Full benchmark evaluation across suites (20 ep/task). Orig = original LIBERO tasks; Ext = our extended tasks.}
\label{tab:benchmark_full}
\small
\begin{tabular}{l cc cc cc c cc}
\toprule
 & \multicolumn{2}{c}{Spatial} & \multicolumn{2}{c}{Goal} & \multicolumn{2}{c}{Object} & Libero-10 & \multicolumn{2}{c}{Total} \\
\cmidrule(lr){2-3} \cmidrule(lr){4-5} \cmidrule(lr){6-7} \cmidrule(lr){8-8} \cmidrule(lr){9-10}
Method & Orig & Ext & Orig & Ext & Orig & Ext & Orig & Orig & Ext \\
\midrule
$\pi$0.5 & 97.0 & 5.9 & 97.0 & 30.0 & 100.0 & 37.7 & 81.0 & 93.8 & 21.4 \\
$\pi$0 & 47.0 & 3.6 & 63.0 & 0.0 & 43.0 & 18.6 & 40.0 & 48.3 & 8.6 \\
$\pi$0-FAST & 65.0 & 1.5 & 37.8 & 1.2 & 61.0 & 5.0 & 26.0 & 47.5 & 2.7 \\
SmolVLA & 17.0 & 3.2 & 44.0 & 0.0 & 50.0 & 13.2 & 41.0 & 38.0 & 6.4 \\
$\pi$0.5-Ours (45-task) & 95.0 & \textbf{10.2} & 85.0 & 27.2 & 100.0 & 37.0 & 78.0 & 89.5 & \textbf{22.8} \\
$\pi$0.5-Ours (56-task) & 97.0 & 7.1 & 77.0 & 26.1 & 98.0 & 35.0 & 70.0 & 85.5 & 20.4 \\
\bottomrule
\end{tabular}
\end{table*}
\begin{table*}[ht]
\centering
\caption{Extended task evaluation by perturbation dimension (20 ep/task).}
\label{tab:benchmark_dimension}
\small
\begin{tabular}{lccccc}
\toprule
Method & Ch.\ Obj.\ (38) & Ch.\ Tgt.\ (13) & Sp.\ Desc.\ (5) & Drawer (3) & Total (59) \\
\midrule
$\pi$0.5 & 29.3 & 0.0 & 11.0 & 31.7 & 21.4 \\
$\pi$0 & 10.8 & 0.0 & 20.0 & 0.0 & 8.6 \\
$\pi$0-FAST & 3.1 & 2.3 & 2.5 & 0.0 & 2.7 \\
SmolVLA & 7.6 & 0.0 & 8.0 & 16.7 & 6.4 \\
$\pi$0.5-Ours (45-task) & 28.4 & \textbf{6.2} & \textbf{25.0} & 20.0 & \textbf{22.8} \\
$\pi$0.5-Ours (56-task) & 27.5 & 5.0 & 17.0 & 3.3 & 20.4 \\
\bottomrule
\end{tabular}
\end{table*}

\subsection{Analysis}

\textbf{Finding 1: Dilution effect.}
Comparing 6-task and 45-task results, the dilution effect is evident. 6-task achieves 28\% on 5 spatial extended tasks, while 45-task achieves only 4\% on the same tasks. Adding large amounts of official training data dilutes the effect of extended task training.

\textbf{Finding 2: Why 6-task improves while 16-task decreases.}
6-task improves from 0\% to 28\%, while 16-task drops from 26.2\% to 6.2\%. This apparent contradiction has the following explanation:

\begin{itemize}
    \item \textbf{6-task}: These 5 libero\_spatial extended tasks have no pattern matching with official tasks, so the official checkpoint baseline is 0\%. Our extended data training brings genuine improvement.
    \item \textbf{16-task}: Among these 16 cross-suite extended tasks, some (especially object and goal suites) have pattern matching with official tasks, resulting in a 26.2\% baseline. When training only on 16 extended tasks without official data, the model loses this pattern matching capability, causing the decrease.
\end{itemize}

\textbf{Finding 3: Double-edged sword of scaling.}
16-task achieves only 6.2\%, while 56-task reaches 27.5\%. On the surface, adding official data helps. However, on the spatial dimension, 56-task achieves only 6.7\%. This indicates that official data helps pattern matching on familiar tasks or extended tasks similar to official ones, but cannot transfer language understanding to novel variations.

\textbf{Finding 4: Generalization comparison of 45-task vs 56-task.}
On the full benchmark (59 extended tasks), 45-task achieves 10.2\% on Spatial, outperforming 56-task's 7.1\%. Overall, 45-task (22.8\%) also slightly outperforms 56-task (20.4\%). This further confirms the dilution hypothesis.

\textbf{Finding 5: Fundamental challenge of VLA language understanding.}
Single-task fine-tuning can improve from 3.75\% to 90\%, demonstrating the model's learning capability. However, as task scale increases, even with targeted same-scene multi-task data, the model still struggles to establish generalizable language understanding. This reveals a fundamental limitation of current VLA models.

\section{Discussion}

\textbf{Why does performance degrade with scale?}
The progressive pattern---90\% (single-task) $\rightarrow$ 28\% (6-task) $\rightarrow$ 6.2\% (16-task)---reveals the difficulty of multi-task learning.

At single-task scale, the model achieves 90\% through memorization, but this is not genuine language understanding. At 6-task scale, same-scene multi-task training shows partial effectiveness: Spatial Description (60\%) is more successful than Change Target (10--40\%) or Change Object (10--20\%), indicating that difficulty increases as the semantic distance between extended tasks and pre-training patterns increases.

At larger scales, the problem becomes more severe. 16-task (extended tasks only) achieves only 6\%---once extended tasks increase to 16, the model struggles to learn. 56-task shows no substantial difference from the official checkpoint on extended tasks, and actually decreases on official tasks (93.8\% $\rightarrow$ 85.5\%). This indicates that merely adding same-scene multi-task data has limited effectiveness in addressing language understanding. To truly improve understanding under large-scale, diverse semantic perturbations, more effective model architectures and training strategies are needed in conjunction with further increasing semantically varied data.

\textbf{What does cross-model comparison reveal?}
$\pi$0-FAST achieves only 2.7\% on extended tasks, far below $\pi$0.5's 21.4\%. This reveals several insights: (1) Model architecture significantly impacts language understanding---FAST's action chunking design may rely more heavily on visual pattern matching; (2) Our benchmark can serve as a diagnostic tool to measure the language understanding capabilities of different models; (3) When selecting VLA models for deployment, one should not only consider accuracy on official tasks, but also language generalization ability.

\textbf{Why is Change Target the most difficult?}
All 13 tasks in the Change Target dimension achieve 0\% on the baseline, and only reach 5--6\% even after training. This indicates that the model almost completely ignores linguistic descriptions of target locations. In contrast, Change Object (29\%) and Drawer Action (32\%) perform better, possibly because these dimensions are closer to patterns in the pre-training data. This finding has implications for architecture design: future VLA models need dedicated mechanisms to handle spatial relation descriptions.

\textbf{Limitations.}
Our four semantic dimensions do not exhaust linguistic variation. More complex language structures such as temporal ordering (``first pick up the bowl, then open the drawer) or negation (``do not place on the plate) remain unexplored. Additionally, the current work focuses on simulation environments; transfer to real robot scenarios requires further research.

\textbf{Future directions.}
Per-suite data balancing and curriculum learning that prioritizes language-gap tasks are natural next steps. Our data-centric approach is complementary to architectural solutions (BayesVLA~\cite{ke2025bayesvla}, LangForce~\cite{wang2026langforce}, Zhan et al.~\cite{chen2026rss})---combining modality-rebalancing architectures with data that requires language reliance is a promising direction. Scaling task generation through automated demonstration synthesis~\cite{mandlekar2023mimicgen} or LLM-assisted instruction synthesis could extend the taxonomy to additional dimensions.

\section{Conclusion}

We introduce a semantic perturbation taxonomy and LangGap, a 99-task benchmark where identical visual states make language instructions the only discriminative signal. Applying the taxonomy to $\pi$0.5 reveals differential grounding failures---0\% on the Change Target dimension vs.\ 29.3\% on the Change Object dimension---which cannot be finely analyzed using existing aggregate metrics without perturbation-based categorization. Systematic progressive experiments demonstrate that same-scene multi-task data can effectively narrow the gap at relatively small scales, but becomes diluted under joint training. This confirms that merely adding multi-semantic-perturbation data under the same visual layout has limited effectiveness in addressing language understanding. To truly improve scalable and diverse language understanding, stronger model architectures and training strategies are needed in conjunction with further increasing language diversity. Our approach is complementary to architectural solutions, and both represent important directions for improving language understanding in VLA models that can mutually reinforce each other. Therefore, the benchmark and data collection training paradigm proposed in this paper can serve as a long-term resource for evaluating the genuine diverse language understanding capabilities of VLA models.

\bibliographystyle{IEEEtran}
\bibliography{references}

\appendix
\section{Detailed Diagnostic Results}

Tables~\ref{tab:diag_target}--\ref{tab:diag_drawer} provide per-task breakdowns of the $\pi$0.5 diagnostic evaluation (Section~III-B), organized by perturbation dimension.

\vfill\null\newpage

\centering
\footnotesize
\captionof{table}{Change Target: $\pi$0.5 diagnostic detail (20 ep/task).}
\label{tab:diag_target}
\begin{tabular}{llr}
\toprule
Suite & Perturbation & Success \\
\midrule
spatial & plate $\rightarrow$ stove (4 tasks) & 0/80 \\
spatial & plate $\rightarrow$ cabinet (4 tasks) & 0/80 \\
spatial & stove $\rightarrow$ plate (2 tasks) & 0/40 \\
spatial & stove $\rightarrow$ cabinet (2 tasks) & 0/40 \\
goal & plate $\rightarrow$ stove (1 task) & 0/20 \\
\midrule
\multicolumn{2}{l}{\textit{Total (13 tasks)}} & \textbf{0/260 (0\%)} \\
\bottomrule
\end{tabular}

\vspace{6pt}

\captionof{table}{Change Object: $\pi$0.5 diagnostic detail (20 ep/task).}
\label{tab:diag_object}
\begin{tabular}{llr}
\toprule
Suite & Perturbation & Success \\
\midrule
spatial & bowl $\rightarrow$ ramekin / cookie (8) & 3/160 (1.9\%) \\
goal & various (8 tasks) & 54/160 (33.8\%) \\
object & various (22 tasks) & 166/440 (37.7\%) \\
\midrule
\multicolumn{2}{l}{\textit{Total (38 tasks)}} & \textbf{223/760 (29.3\%)} \\
\bottomrule
\end{tabular}

\vspace{6pt}

\captionof{table}{Spatial Description: $\pi$0.5 diagnostic detail (20 ep/task).}
\label{tab:diag_spatial}
\begin{tabular}{llr}
\toprule
Suite & Perturbation & Success \\
\midrule
spatial & task0\_bowl2 & 1/20 (5\%) \\
spatial & task2\_bowl2 & 0/20 (0\%) \\
spatial & task3\_bowl2 & 9/20 (45\%) \\
spatial & task4\_bowl2 & 0/20 (0\%) \\
spatial & task8\_bowl2 & 1/20 (5\%) \\
\midrule
\multicolumn{2}{l}{\textit{Total (5 tasks)}} & \textbf{11/100 (11.0\%)} \\
\bottomrule
\end{tabular}

\vspace{6pt}

\captionof{table}{Drawer Action: $\pi$0.5 diagnostic detail (20 ep/task).}
\label{tab:diag_drawer}
\begin{tabular}{llr}
\toprule
Suite & Perturbation & Success \\
\midrule
spatial & open top drawer & 6/20 (30\%) \\
spatial & open middle drawer & 13/20 (65\%) \\
spatial & open bottom drawer & 0/20 (0\%) \\
\midrule
\multicolumn{2}{l}{\textit{Total (3 tasks)}} & \textbf{19/60 (31.7\%)} \\
\bottomrule
\end{tabular}

\end{document}